\documentclass[letter]{article}
\usepackage[totalwidth=420pt,totalheight=625pt]{geometry}
\usepackage{authblk}

\date{Last updated: Feb 19, 2021}

\usepackage[USenglish]{babel}
\usepackage{graphicx}
\graphicspath{{1_figs/}}
\usepackage{booktabs}
\usepackage{array}
\newcolumntype{H}{>{\setbox0=\hbox\bgroup}c<{\egroup}@{}}
\usepackage{multirow}
\usepackage{diagbox}
\usepackage{xspace}

\newcommand{\nop}[1]{}

\usepackage{tikz}
\usetikzlibrary{tikzmark}

\usepackage{xcolor}
\colorlet{MyBlue}{blue!50!black!100!}
\colorlet{MyRed}{red!50!black!100!}

\usepackage[hyphens,spaces]{url}
\usepackage{hyperref,xcolor}  
\usepackage{accsupp}

\newcommand{\set}[1]{\textit{#1}}

\newcommand{\solver}[1]{\texttt{#1}}
\newcommand{\SAThy}{SAT-}
\newcommand{\teamA}{SW\xspace}
\newcommand{\teamB}{HW\xspace}

\newcommand{\sFIXED}[1]{#1}
\newenvironment{FIXED}
  {%
    \begingroup%
  }
  {%
    \endgroup
  }%

\def\citexds#1,#2\relax{\citeauthor{#1}~\cite{#1,#2}}

\begin{document}

\title{A Time Leap Challenge for \SAThy Solving%
}

\author[1]{Johannes K. Fichte}%
\author[2]{Markus Hecher}%
\author[3]{Stefan Szeider}%

\affil[1]{\small Link\"oping University, Sweden\\
    \texttt{johannes.fichte@liu.se}}
\affil[2]{Massachusetts Institute of Technology, United States\\
    \texttt{hecher@mit.edu}}
\affil[3]{TU Wien, Austria\\
    \texttt{sz@ac.tuwien.ac.at}}
  \maketitle     

  \begin{abstract}
    We compare the impact of hardware advancement and algorithm
    advancement for \SAThy solving over the last two decades. In
    particular, we compare  20-year-old \SAThy solvers
    on new computer hardware with 
    modern  \SAThy solvers on 20-year-old hardware. Our findings
    show that the progress on the algorithmic side has at least as much
    impact as the progress on the hardware side.
    
  \end{abstract}
\section{Introduction}
The last decades have brought enormous technological progress and
innovation. Two main factors that are undoubtedly key to this
development are (i)~\emph{hardware advancement} and
(ii)~\emph{algorithm advancement}. Moore's Law, the prediction made by
Gordon Moore in 1965 \cite{Moore65}, that the number of components per
integrated circuit doubles every year, has shown to be astonishingly
accurate for several decades. Given such an exponential improvement on
the hardware side, one is tempted to overlook the progress on the
algorithmic side.
 
This paper aims to compare the impact of hardware advancement and
algorithm advancement based on a genuine problem, the propositional
satisfiability problem (SAT). This problem is well-suited for such a
comparison since it is one of the first problems for which progress in
solving has been measured regularly through
competitions~\cite{JarvisaloBRS12}. Also, a standardized instance
format has been established very early.  By focusing on this problem,
the comparison allows us to fathom the SAT and CP community's
contribution to the overall progress.

Of course, the advancements in hardware and algorithms cannot be
separated entirely.  Targeted algorithm engineering can use
new hardware
features~\cite{BornebuschWilleDrechsler17a,ChuHarwoodStuckey09a,FichteMantheySchidler20a,JarvisaloHeuleBiere12}, 
and hardware development can be guided by the specific demands of
modern algorithms.
We are well aware that this can quickly end up in comparing apples and
oranges.
Nevertheless, we think that by carefully setting up the experiment and
choosing hardware and algorithms, it still allows us to draw some
conclusions on the  individual components' impact.
  
We base the general setup of the comparison on a \emph{Time Leap
  Challenge}, where virtual teams compete. Team~\teamA uses new
solvers on old hardware; Team~\teamB uses old solvers on new hardware.
The time between ``old'' and ``new'' spans about two decades. Which
team can solve more instances? Depending on the outcome, one can
compare the impact of hardware advancement and algorithm
advancement. The idea for this time leap challenge for \SAThy solvers
was inspired by a thought experiment on algorithms in mixed-integer
linear programming (MILP), suggested by Sebastian
Stiller~\cite{Stiller15a}.
\sFIXED{%
  Furthermore, we received inspirations from work by
  Brixby~\cite{Bixby02,Bixby12}, who carried out experimental work
  comparing old and new solvers limited to one, at the time, modern
  computer in the area of mixed integer linear programming.
}

In the early 1990s, the dominant complete method for \SAThy solving was
the \emph{DPLL Algorithm} (Davis-Putnam-Logemann-Loveland
\cite{DavisPutnam60,DavisLogemannLoveland62}), which combines
backtracking search with Boolean constraint
propagation~\cite{ZabihMcallester88}. However, in the late 1990s, the
\emph{CDCL Solvers} (Conflict-Driven Clause Learning) took over. They
extended the basic DPLL framework with new methods, including clause
learning~\cite{MarquessilvaSakallah96}, lazy data structures like
watched literals~\cite{MoskewiczEtAl01},
backjumping~\cite{MarquessilvaSakallah96,MoskewiczEtAl01}, and dynamic
branching heuristics~\cite{MoskewiczEtAl01}; the combination of these
methods resulted in a significant performance boost, often referred to
as the ``CDCL Revolution''. 
Although the CDCL paradigm is still predominating today's \SAThy solving,
there have been several significant improvements made over the last
two decades, including efficient preprocessing \cite{EenBiere05} and
inprocessing \cite{JarvisaloHeuleBiere12}, aggressive clause
deletion~\cite{AudemardSimon09}, fast restarts
\cite{LubySinclairZuckermann93}, lightweight component caching
\cite{PipatsrisawatDarwiche07}, implication queue
sorting~\cite{LewisSchubertBecker05}, and new branching
heuristics~\cite{LiangGaneschPaupartCzarnecki16}.  

\subsection{Experimental Setting}
For our Time Leap Challenge, Team~\teamB (old solvers on new hardware) is
composed of the solvers \solver{Jerusat} (2003), \solver{siege\_v3} (2003),
and \solver{Forklift} (2003) running on a computer from 2019 with an
Intel Xeon Silver 4112 CPU at 2.60GHz base frequency and 128GB RAM.
Team~\teamA (new solvers on old hardware) is composed of the solvers
\solver{MapleSat19} (2019), \solver{CaDiCal} (2019), and
\solver{Glucose} (2016) running on a computer from 1999 with a
Pentium~III processor at 467MHz frequency and 1.5GB RAM.
An essential question for setting up the experiment was the choice of
a suitable set of benchmark instances. On the one hand, the instances
should not be too challenging so that they are not entirely out of
reach for old solvers or old hardware; on the other hand, the
instances should still be challenging enough to provide interesting
results. We settled on the benchmark
set~\textit{set-asp-gauss}~\cite{HoosKaufmannSchaub13a} that provides a
reasonably good compromise, as it contains a large variety of
instances, tailors adapted instance hardness, is free of duplicates,
reproducible, and publicly available.  We used a timeout of 900
seconds, which is the default for SAT competitions.
Right in the beginning, we state a clear disclaimer. While a
theoretical challenge is easy to design, a practical comparison can
rarely be comprehensive and complete. About 20 years of evolution
increases the practical search space by orders. There are many
possibilities to combine hardware, software, benchmarks, and
solvers. Particularly, there might be solvers that are still
available, but we missed during our research.  Still, we provide a
clear guideline on how we selected the teams and provide extensive
details beyond. Our results are reproducible in the setting, and the
conclusions provide a general idea. However, the ideas might not
generalize to conclusions over other benchmark sets or solvers we
might have missed. However, this is a usual situation in many
experiments with combinatorial solving as there is no good theoretical
understanding of the practical effects~\cite{Nordstrom15a}.  Still, we
aimed to put the concept of a thought time leap challenge from
literature in popular science into a practical scientific context.
\subsection{Results}
Table~\ref{table:summary} gives a summary of our results (we provide
more details in Section~\ref{sec:results}).  We see that both teams
perform in a similar range with a slight advantage for Team~\teamA.

\begin{FIXED}
\subsection{Prior Work}
This paper is an extended and updated version of a paper that appeared
in the proceedings of the 26th International Conference on Principles
and Practice of Constraint Programming (CP
2020)~\cite{FichteHecherSzeider20}.
The present paper extends the preliminary work by %
additional experimental work, including solvers from the SAT heritage
project launched in Summer 2020~\cite{AudemardPauleveSimon20a}.
Furthermore, in contrast to our previous
work~\cite{FichteHecherSzeider20}, we check whether the solvers output
the correct decision~\cite{Biere20}. We limit ourselves to a check, since the old solvers do not include techniques for proof checking unsatisfiable instances. 
Finally, we include the results of the solvers on a more recent CPU
generation.

\end{FIXED}

\begin{table}[t]

\centering
\[
\begin{array}{@{}r@{\hskip 0.5em}@{\hskip 0.5em}c@{\hskip 0.5em}@{\hskip 0.5em}c@{\hskip 0.5em}@{\hskip 0.5em}c@{\hskip 0.5em}@{\hskip 0.5em}c@{\hskip 0.5em}@{\hskip 0.5em}c@{\hskip 0.5em}@{\hskip 0.5em}c@{}}
  \toprule
   & \solver{Jerusat} & \solver{siege\_v3}  & \solver{Forklift}  & \solver{Glucose}  & \solver{CaDiCal}  & \solver{MapleSat19}  \\
   & (2003) & (2003) & (2003) & (2016) & (2019) & (2019) \\[1.8em]
    \hbox{old~HW (1999)} & 38 & 43 & 43 & \tikzmark{startS}105 & 100 & 72\tikzmark{endS} \\[1.5em]
    \hbox{new~HW (2019)} &\tikzmark{startH} 71 & 96 & 105 \tikzmark{endH}& 188 & 184 & 187 \\[0.5em]
  \bottomrule
\end{array}
\]
\begin{tikzpicture}[remember picture,overlay]
\foreach \Val in {S,H}
{
\draw[rounded corners,black,thick]
  ([shift={(-1.5\tabcolsep,-1.5ex)}]pic cs:start\Val) 
  node[above right,xshift=2.6em,yshift=1.4em]{\small \sffamily Team \Val{}W}
    rectangle 
  ([shift={(1.5\tabcolsep,2ex)}]pic cs:end\Val);
}
\end{tikzpicture}
\caption{Summary of experimental results}
\label{table:summary}
\end{table}

\nocite{LiangGaneschPaupartCzarnecki16,LubySinclairZuckermann93,EenBiere05,
LewisSchubertBecker05,PipatsrisawatDarwiche07,AudemardSimon09, GomesSelmanKautz98,ZabihMcallester88,JarvisaloHeuleBiere12}

\subsection{Related Work}
Knuth~\cite{Knuth15a} provides an overview of various aspects of \SAThy
solving, including commented implementations of algorithms from
several epochs of \SAThy solving.
His implementations assemble a DPLL solver (\solver{SAT10}), a DPLL
look-ahead solver (\solver{SAT11}), and a CDCL solver
(\solver{SAT13}), as well as a preprocessor (\solver{SAT12}). Since
all these solvers are implemented uniformly, without special
implementation or hardware tricks, they provide an excellent
comparison of the algorithmic advancement of solver techniques. We
therefore included, for comparison, the results of Knuth's solvers on
the same benchmark set and hardware platform as the time leap
challenge. 
Mitchell~\cite{Mitchell05} provides an overview of techniques,
  implementations, and algorithmic advances of the year 2005 and
  looking back for 15 years.
  He already mentioned that the success of \SAThy solving is due to
  three factors: improved algorithms, improved implementation
  techniques, and increased machine capacity. However, Mitchell's work
  does not provide evaluations on any actual practical effects at the
  time.
  Kohlhase~\cite{Kohlhase19a} recently published work on collecting
  and preserving the comparability of old theorem provers to preserve
  cultural artifacts and history in Artificial
  Intelligence.
  For an overview on the technique of CDCL-based solvers we refer the
  reader to introductory literature such as a chapter in the Handbook
  of Knowledge Representation~\cite{GomesKautzSabharwalSelman08},
  chapters on the history of modern \SAThy
  solving~\cite{FrancoMartin09a}, and
  CDCL-solvers~\cite{Marques-SilvaLynceMalik09a} in the Handbook of
  Satisfiability~\cite{BiereHeuleMaarenWalsh09}.
Katebi, Sakallah, and
Marques-Silva~\cite{KatebiSakallahSilva11a,SakallahSilva11a}
considered various techniques of modern \SAThy solvers from an
empirical viewpoint. They designed experiments to evaluate factors and
the aggregation of different \SAThy enhancements that contribute to
today's practical success of modern solvers.
Works on targeted algorithm engineering for \SAThy solvers are
extensive. Just to name a few examples, there is work on exploiting
features such as optimizing memory footprints for the
architecture~\cite{BornebuschWilleDrechsler17a}, on implementing
cache-aware~\cite{ChuHarwoodStuckey09a}, on using huge
pages~\cite{FichteMantheySchidler20a}, on how to benefit from
parallel solving~\cite{IserBalyoSinz19a} or employing
inprocessing. Inprocessing particularly takes advantage of modern
hardware as one can execute a lot more instructions on a modern CPU
than accessing bytes on
memory~\cite{HennessyPatterson11a,MahapatraVenkatrao99a}.
  Very recently,
  Audemard, Paulevé, and Simon~\cite{AudemardPauleveSimon20a}
  published a heritage system for SAT solvers. It allows for
  compiling, archiving, and running almost all released SAT solvers
  and is based on Docker, GitHub, and Zenodo. While they aim for
  archivability, our work provides an actual experiment incorporating
  soft- and hardware advances. We hope that their system allows for
  long term preservation and, if there is no major change in
  computer architecture, one can repeat our time leap challenge
  in another decade.

\section{The Arena: Designing the Time Leap Challenge}
To run a proper challenge, we design an arena by selecting
from standard benchmark sets and several contestants out of a vast
space of possibilities.
We aim for the oldest reasonable hardware on which we can still run
modern benchmark sets and solvers. In turn, this requires setting up a
modern operating system on old hardware.
To make it a time leap challenge, we are interested in
solvers and hardware from similar generations, in other words in a preferably small
time frame from which both originate.
The physical effort restricts us to consider only two time frames in
the following. We take modern hardware and solvers from 2019 and old
hardware from around 2000 and solvers from 2001/2002.
Following academic ideas by Stallman~\cite{Stallman85a}, we focus on
benchmark sets and solvers that are publicly available.
Throughout the experimental work, we follow standard guidelines for
benchmarking~\cite{KouweAndriesseBos18a}.
In the course of this section, we elaborate on 
various technical and organizational obstacles. Setting up a time leap
challenge is also somewhat of an archaeological challenge.

In theory, a variety of competitions have been organized in the
past. The competition results give a broad picture of benchmark
instances and solvers. Old hardware and operating systems should still
be widely available.
In practice, neither open source, nor version control systems, nor
public platforms to host software projects such as
SourceForge~\cite{sourceforge}, bitbucket, github, or gitlab, were
popular in the community around the millennium. Publicly funded data
libraries such as Zenodo~\cite{Nielsen19a} were also established much
later.
While the culture of storing text in libraries dates back to
Alexandria and the first librarian Zenodotus in 280~BC, searching for
datasets and source codes from 20 years ago feels like digging through
a burnt library. 
Enthusiasts maintained datasets and source codes from early competitions.
Sometimes source codes were kept as a
secret~\cite{GoldbergNovikov03}. Some links redirect to grabbed
domains, or people moved and with them, the webpages. Sometimes binaries
show up from private collections or the Internet
Archive~\cite{Kahle20a}. However, it turned out that they do not
run, as libraries on which they depend, do not run on modern Linux or
Unix distributions.
\smallskip
Below we report and explain details of the selection process.

\paragraph{Instance Format}
Johnson and Trick suggested a uniform input format description in
1993, which is still used as the standard for SAT input
instances~\cite{JohnsonTrick93a}. The standardized input format and
backward compatibility substantially simplified our selection
process.

\subsection{Selecting a Suitable Benchmark Set}
Our focus on selecting a benchmark set is to consider a larger benchmark
set, say of a cardinality ranging from 100 to~300. We are interested
in a safe and stable choice of instances since benchmarks run a wide
variety of experiments with preferably more than 10 solvers resulting
in months of running time.
Hence, we push to a reasonable state-of-the-art benchmark setting.
We prefer instances that (i)~are publicly available, (ii)~contain a
good selection of domains, including an industrial background, random,
and combinatorial instances, and (iii) highlight differences for
modern solvers.
We summarize runtime and number of solved instances during our
instance selection process in Table~\ref{tab:selection}.
For an initial selection, we ran instances only with the solver
\solver{Glucose}~\cite{glucose421}, which showed robust performance on
many earlier experimental works that we carried out.

\begin{table}%
\centering
\def\arraystretch{1.2}%
\begin{tabular}{@{\hskip 1em}c@{\hskip 1em}@{\hskip 1em}c@{\hskip 1em}@{\hskip 1em}r@{\hskip 1em}r@{\hskip 1em}r@{\hskip 1em}r@{\hskip 1em}r}
\toprule
benchmark & solver &  \# & TO & ERR & $t[h]$ & avg[s]\\
\midrule
 DIMACS2 & \solver{Glucose} & 225 & 15 & 1 & 0.34 & 5.46\\
SATLIB &\solver{Glucose} &  43892 & 15 & 6399 & 4.45 & 0.36\\
set-asp-gauss & \solver{Glucose} & 189 & 11 & 0 & 4.50 & 85.71 \\
\bottomrule
\end{tabular}
\medskip
\caption{Runtime of a modern solver and modern hardware on selected benchmark sets. 
\# refers to the number of solved instances, 
TO refers to the number of instances on which the solver timed out, 
ERR refers to the number of instances on which the solver found an input error,
$t[h]$ refers to the total running time on the solved instances in hours,
avg[s] refers to the average running time of an instance.}
\label{tab:selection}
\end{table}

\paragraph{Available Instances}
The first available benchmark instances \set{DIMACS-2} date back to
1992 and the 2nd \mbox{DIMACS} Challenge 1992--1993 on NP-hard
problems, which also considered SAT as a
problem~\cite{TrickChvatalCook93a}.
The 241 instances are still well maintained and
downloadable~\cite{TrickEtAl92}.
Note that the 1st SAT competition already took place in
1992~\cite{BuningBuro93a}. However, the instances are not publicly
available.
Over time researchers collected benchmarks such as
\set{SATLIB}~\cite{Hoos00a}, which count more than 50,000 instances in
total. The instances are still available on an old webpage by the
collector~\cite{Hoos00a}.
A subset of these instances was also used for the  SAT
Competition 2002. However, those instances are not available from the SAT
Competition website due to an abandoned domain.
Instances from one of the annual SAT competitions from
2002 to 2019~\cite{sat_competition20}
follow stricter rules, and detailed reports are
available~\cite{JarvisaloBerreRoussel12a}.
There are plenty of tracks, thousands of instances, and many of the
more modern instances are enormous in size.
A popular benchmark set with various instances from SAT
competitions until 2013 and various fields is the benchmark set
\set{set-asp-gauss}~\cite{HoosKaufmannSchaub13a}. 
The set is a composition of representative benchmarks from a
  variety of sources. It has been widely used as a robust selection
  for tuning solvers in the past and was obtained by classifying the
  practical hardness of the instances from the SAT Competition 2009
  and SAT Challenge 2012 and then selecting instances by sampling with
  the Gaussian probability distribution~\cite{HoosKaufmannSchaub13a}.

\paragraph{Initial Evaluations}
In order to gather initial insights, we ran all available solvers on
our cluster.
The hardware for the benchmark selection process consisted of a
cluster of RHEL~7.7 Linux machines equipped with two Intel Xeon
E5-2680v3 CPUs of 12 physical cores each running at 2.50GHz, which we
enforced by performance governors.  The machines are equipped with
64GB main memory of which 60.5GB are freely available to programs.  We
compare wall clock time and number of timeouts.  However, we avoid IO
access on the CPU solvers whenever possible,~i.e., we load instances
into the RAM before we start solving. We run four solvers on one node
at most, set a timeout of 900 seconds, and limit available RAM to~8GB
per instance and solver.
We summarize our initial evaluation of the early benchmark sets in
Table~\ref{tab:selection}.
The DIMACS-2 instances turned out to be very easy for modern solvers.
For example, the solver Glucose solved almost all instances within
less than one second, only five large instances (par32-X.cnf) of a parity
learning problem remained unsolved within 900 seconds.
The SATLIB instances are more challenging but still fairly easy for
modern solvers.
The SAT Competition 2002--2019 instances provide a broad
selection. Since the results are still publicly available, we
refrained from rerunning these sets.
The runtime results on the benchmark set \set{set-asp-gauss} revealed
that modern solvers can  solve many instances. However, the
instances are still challenging as the overall runtimes are reasonably
long. Old solvers are still able to solve plenty of instances on 
modern hardware. The benchmark set consists of 200 instances in total.

\paragraph{Decision} 
After running the instances, we picked one existing benchmark set.
Since the set \set{DIMACS-2} contains almost only easy instances, we
rejected the set right away. While the \set{SATLIB} instances contain
mainly easy instances, they are not very challenging for modern
solvers. Further, the contained benchmarks have a strong bias towards
handcrafted and random instances.
The SAT 2002--2019 instances contain very interesting sets. However,
some of the more modern instances are very large, and we
figured that it is impossible to transfer and run the instances on old
hardware.
After reviewing the initial results and sampling memory requirements
from earlier SAT competitions, we decided to use the benchmark set
\set{set-asp-gauss}~\cite{HoosKaufmannSchaub13a}, which provides a
reasonably good compromise. It contains a large variety of instances,
tailors adapted instance hardness, is free of duplicates,
reproducible, and publicly available.

\subsection{Selecting Solvers}
In the following section, we describe the selection process of \SAThy
solvers for our challenge.
To foster reproducibility and favor open-source, we focus on
publicly available solvers (binary or source code).
Note that modern \SAThy solving also includes various parallel
algorithms. Due to the unavailability of wide parallel computation on old
hardware, we restrict ourselves to sequential solvers.
Further, we consider only solvers that are, vaguely speaking, 
descendants of the DPLL~\cite{DavisPutnam60,DavisLogemannLoveland62}
algorithm,~i.e., CDCL. 
These solvers are often referred to as solvers implementing complete
and systematic search.  However, restarts and deletion might affect
completeness under certain conditions in
practice~\cite{Marques-SilvaLynceMalik09a}.
To our knowledge, CDCL-based solvers with various additional
techniques on top, which even extend the underlying proof system, are
still the most prevailing paradigm for \SAThy solvers. However, today,
some solvers use strong proof techniques such as the division rule in
cutting planes~\cite{ElffersNordstrom18a,GochtNordstromYehudayoff19a}
or Gaussian Elimination~\cite{Soos10a,Soos18a}.

\paragraph{Researching for Solvers}
The 1st SAT Competition~\cite{BuningBuro93a} and 2nd \mbox{DIMACS}
Challenge~\cite{TrickChvatalCook93a} took place around 1992. However,
no online resources on detailed solvers or source codes are available.
The earliest public collection of solvers, which is still available
online, 
is the SATLIB Solver Collection~\cite{Hoos00b}.
The collection contains implementations on DPLL-based implementations
as well as stochastic local search solvers.
DPLL-based Implementations in the collection are
\solver{Grasp}~\cite{MarquessilvaSakallah96}, 
\solver{NTAB}~\cite{CrawfordAuton93a},
\solver{POSIT}~\cite{Freeman95a},
various versions of
\solver{REL\_SAT}~\cite{BayardoSchrag97a,BayardoPehoushek00}, which are also
available on github~\cite{Bayardo07}, 
two versions of \solver{SATO}~\cite{Zhang97a}, and
four versions of \solver{Satz}~\cite{LiAnbulagan97a}.
Further, we asked colleagues for the source code of old solvers and
received an even older version of \solver{Grasp}
from~1996~\cite{Marques-Silva20a}.
The era of CDCL solvers started in 2001~\cite{MoskewiczEtAl01}. There, 
successful solvers such as \solver{BerkMin}~\cite{GoldbergNovikov03},
\solver{siege}~\cite{Ryan03a}, and
\solver{zChaff}~\cite{FuMahajanMalik04a} materialized.
\solver{Siege}~\cite{Ryan03a} is publicly available with binaries in three
versions from 2003 to 2004.  We contacted colleagues on the source
code of \solver{siege}, but the author retired, and the sources seem to be lost.
For \solver{zChaff}~\cite{FuMahajanMalik04a} even the source code is publicly
available in four versions from 2001 to 2007.
Binaries of \solver{BerkMin} showed up in a backup of experiments on \SAThy
solvers from earlier works. We contacted the authors on source codes
but received no answer.
\begin{FIXED}
  Audemard, Paulev\'e, and Simon~\cite{AudemardPauleveSimon20a} very
  recently started the SAT Heritage project, which provides software
  heritage~\cite{SoftwareHeritage20} for the SAT community.
  The SAT Heritage project contains binaries or source codes for
  solvers starting from the early SAT competitions in 2002.%
  \footnote{%
    Note that we did not run the solver within SAT Heritage, but
    extracted relevant solvers. Audemard, Paulev\'e, and Simon provide
    an easy option on that account. After installing Docker and then
    SAT Heritage by running \texttt{pip install -U satex}, one can
    simply extract the solver with the command \texttt{satex extract
      berkmin561:2003 myfolder}.
  }%
  Among the binaries, we also found a working binary of
  \solver{BerkMin} and its successor \solver{Forklift} from 2003.
\end{FIXED}%
A famous solver in the \SAThy solvers line is \solver{MiniSat}, which is
available online
in various
versions~\cite{EenSorensson08,EenSorensson04a,SorenssonEen05}.
The development of \solver{MiniSat} started around 2003~\cite{EenSorensson04a}
intending to create a compact, readable, and efficient solver for
the community. The earliest version online is from 2005, and the most
known and very popular version~2.2 from 2008.
Another popular \SAThy solver is \solver{Glucose}~\cite{AudemardSimon12a},
which was developed to aggressively remove clauses that are not
helpful during clause learning of the CDCL procedure. This results in
an incomplete algorithm as keeping learnt clauses is essential for
completeness.  We consider the version \solver{Glucose}
syrup~4.2.1~\cite{glucose421}.
A very popular, successful and recent solver is
\solver{Lingeling}~\cite{Biere17a}, which won several SAT competitions and the
prize on the most innovative solver~\cite{BalyoBiereIser16a} in 2015.
Two medalists of the SAT 2019 Race were
\solver{CaDiCaL}~1.0.3~\cite{Biere19a} and a descendant of the solver
\solver{MapleSAT}~\cite{LiangGaneschPaupartCzarnecki16}, namely
\solver{MapleLCMDistChronoBTDL-v3 (MapleSat19)}~\cite{MapleLCMDistChronoBTDL}.
\begin{FIXED}
  We would like to point the reader at swMATH~\cite{ChraparyRen16},
  an information service for mathematical software that
  allows for researching solvers and publications in which a solver is
  referenced. Indeed, swMATH can be quite helpful when the solver name is
  known.
\end{FIXED}

\paragraph{Testing the Solvers}
To benchmark a solver, we first need to compile it or run the
binary with a modern operating system as there is otherwise no chance
to get the solvers running on modern hardware.
First, we considered all solvers from the SATLIB collection. We were
able to compile and successfully run the solvers \solver{Grasp},
\solver{Relsat}, \solver{Satz}, and \solver{SATO}.
However, we had to modify the source codes and build files so that
they would compile with a modern compiler due to harder
interpretations of language standards in modern compilers.
Since the solvers were originally designed for 32bit Linux, we
compiled the solvers on 32bit Linux and used them late on 64bit Linux
by compatibility layers. While we were also successful in  compiling
solvers on 64bit systems, the 64bit binary would often solve fewer 
instances on the 64bit system or result in many segfaults. We
suspect compatibility issues as either the developers of the old
solvers could not expect  certain datatypes on a future
architecture or implemented sloppy memory management.
All versions of the solver \solver{siege}, which were available as a binary,
still ran on a modern Linux using the 32bit compatibility mode.
We were successful in building all versions of the solver \solver{zChaff}; both
on a 32bit as well as 64bit architecture.
Unfortunately, the solver BerkMin does not run on modern or fairly
recent Linux distribution. It turns out that the binary was compiled
with an old gcc and linked to an old version of the glibc, which we
discovered in an old Red Hat Enterprise Linux, but we were
  unable to integrate it into a modern Linux distribution.
We found that all modern solvers were well maintained and still compiled on
32 and 64bit Linux distributions as well as a 64bit version of NetBSD.

\begin{FIXED}%
\paragraph{Correctness} 
The first SAT competitions revealed that many early SAT solvers were
buggy, produced segfaults, and outputted wrong
results~\cite[Sec. 2.1.2]{SimonBerreHirsch02}.
Since then, solver correctness has been addressed by intensive testing,
including fuzzing~\cite{BrummayerLonsingBiere10}, and extending the
solvers by systems that allow for formal verification of the results
of unsatisfiable instances by using resolution. Proof
logging~\cite{HeuleHuntWetzler13,WetzlerHeuleHunt14a,HeuleBiere16}
moved modern SAT solving far beyond high-performance tools for
combinatorial solving. Proof logging and verification enabled
SAT solvers in mathematical
applications~\cite{HeuleKullmannMarek16,Heule18,BrakensiekHeuleMackey20}.
To incorporate these advancements and make the results
between old solvers comparable, we exclude solvers that produce a
significant number of erroneous results; for more details, see below.
Armin Biere is currently working on providing bugfixes on various
early SAT solvers~\cite{Biere20}.
While we checked results and base our experiment on the original
sources, we believe that it would still be interesting to repeat the
experiment when the improved solvers are publicly available.

\end{FIXED}%

\paragraph{Final Teams}
To have a comparison on theoretical advances in \SAThy
solving between DPLL and CDCL from an abstract perspective and out of
the hand of a programmer, we picked the implementations by Donald
Knuth~\cite{Knuth15a}. The implementations represent particular periods, more precisely, DPLL solver (\solver{SAT10}), a DPLL
look-ahead solver (\solver{SAT11}), and a CDCL solver
(\solver{SAT13}), as well as a preprocessor (\solver{SAT12}).
We still tested the old solvers \solver{Relsat}, \solver{Satz}, and
\solver{SATO}, which resulted in less than~20 solved instances on our
modern hardware for the best solver among them (SATO).
Since it is theoretically well-known that CDCL can be significantly
faster than DPLL~\cite{PipatsrisawatDarwiche09a,Nordstrom15a}, we
already have the solvers by Knuth. There has already been work on the
technological advances of various techniques between techniques in
DPLL and CDCL solvers; we focus on the more modern CDCL solvers for
both teams.
\begin{FIXED}%
Then, we have several solvers available for Team~\teamB from the researched origins and the newly available SAT Heritage project. 
  We focus on the solvers that output results for a considerable
  number of instances.
  We excluded solver \solver{Grasp}, which already implements conflict
  learning, but unfortunately has a bug in the parser yielding
  unpredictable behavior.
  One of the earliest well-known solvers, \solver{zChaff} (2001), also
  produces a total of 9 wrong outputs.
  Similar, the solver \solver{compsat} (2002) outputs 17 wrong
  results and \solver{bmsat} (2003) 8 wrong results.
  We excluded solver \solver{BerkMin561}, as its successor
  \solver{Forklift} performed much better.
  Since the solver \solver{Jerusat} produced only one wrong instance
  on a system from 2013, we decided to include this solver still.
  Hence, there are three solvers left for a team of solvers from
  about~20 years ago (Team~\teamB), namely, \solver{Forklift} (2003),
  \solver{siege} (2003), and \solver{Jerusat} (2003).
\end{FIXED}%
We preferred to include version~3 of the solver \solver{siege} (2003) as it
solved about~12 instances more than version~1 (2001) on our modern
reference hardware.
We discarded \solver{MiniSat} as the youngest of the older solvers.
We picked \solver{CaDiCaL} 1.0.3~\cite{Biere19a} and
\solver{MapleLCMDistChronoBTDL-v3 (MapleSat19)}~\cite{MapleLCMDistChronoBTDL}
for Team~\teamA (new solvers on old hardware) due to their good performance
in the SAT 2019 Race. \solver{MapleSat19} won the SAT 2019 Race, and \solver{CaDiCal}
scored  second place.
Since the slightly older solver \solver{Glucose}
syrup~4.2.1~\cite{glucose421} solved about ten instances more than the
solver \solver{Lingeling} 7d5db72~\cite{Biere17a} on our modern reference
hardware, we decided to pick \solver{Glucose} for our Team~\teamA.

\subsection{Selecting the Environment: Operating System and Compiler}
Since we are interested in comparing the teams ``new solvers on old
hardware'' and ``old solvers on new hardware'', we think that it is only
fair to also include advancements in kernel architecture, compilers,
and operating systems %
for new solvers. Anyway,
it is impossible to obtain ancient Linux or Unix distributions due
to missing source code mirrors. It is not possible to run such
Linux or Unix distributions on modern hardware because of the lack of
modern chipset drivers in ancient kernels.
Due to long term support of hardware, we decided to favor Debian~10
codename buster (July 2019)~\cite{Carter19a} and try NetBSD~9
(Feb. 2020)~\cite{NetBSD-www-team20a} as operating systems.
We ran the experiments on Linux kernel version 4.19.0-8-686-pae.
We use gcc~8.3.0 on Debian and NetBSD.
Our modern hardware at university was equipped with Linux Mint~19
codename Tara, kernel version 4.15.0-91, and gcc compiler
version~7.5.0-3.

\subsection{Selecting the Hardware}
To have a wide variety of hardware, we started to gather old
hardware from friends and colleagues.
We collected ten systems over different generations, namely systems
containing a Pentium~II (1998), a Pentium~III (1999), an Ultra Sparc
IIe (2001), a Pentium~IV (2002), a Pentium IV Prescott (2004), a Core2
Duo (2007), an i5 Nehalem (2009), a Xeon Haswell (2013), a Xeon
Skylake (2017), an i7 Icelake (2019),
\sFIXED{ and an AMD Rome (2020)}.
A colleague prepared a SPARCstation II (1995) and SPARCstation Voyager
(1995) for us.

\paragraph{Technical Restrictions}
The selection of a benchmark set and operating systems restricted the space
of possibilities on the potential old hardware. 
Preferably, we are interested in the oldest possible hardware and the
youngest possible hardware.
In more detail, modern Linux distributions, such as Debian~10, still
support all x86-based (IA-32) i686 processors, including various AMD,
Intel, and VIA processors. However, the i686 architectures limits
experiments to Pentium~II processors (1997) or
later~\cite{JacksonSchwarzMorris19a}. BSD distributions, such as
NetBSD~9, still support the Sparc64 architecture, which in theory
allows for running systems with processors SPARC64 (1995) and
UltraSPARC IIe (1999).
We were able to run NetBSD~9 on a system with the latter CPU 
namely, the Sun Netra X1, from about 2000/2001. Since we only had
access to Linux or Solaris binaries for some solvers, and we were
unable to set Debian~10 or Solaris onto the Netra system in decent
time due to a required setup via serial LOM interface and network
boot, we discarded the Sun system from our final hardware selection.
It is well known that modern operating systems and \SAThy solvers are
very memory-demanding~\cite{FichteMantheySchidler20a} resulting in a
requirement of having at least 1GB of total RAM %
inside the system.
Since the L2 cache controllers of the Pentium~II only allow the use of
512MB of RAM and we could not access a system with a Pentium
Pro processor, our oldest possible system (1999) was a Pentium~III
processor running at 467MHz equipped with 1.5GB RAM. Hence, we picked
this system to run the solvers of Team~\teamA.
While the most modern Intel CPU architecture we had access to was an
i7 Icelake (2019), we decided to prefer the system running a Xeon
Skylake due to the availability of ECC-RAM and the much larger caches,
which are usually beneficial for \SAThy solving.
Still, the modern system with the Xeon Skylate was bought in 2019
for dedicated benchmarking, while the i7 was just a small-form-factor
barebone desktop computer for which we feared that high permanent load
over months might significantly degenerate performance due to
overheating. 
\begin{FIXED}
  Also, we benchmarked solvers on systems equipped with AMD
  Rome CPUs (2020). The behavior was mostly similar, with a few
  instances solved less.
  Still, we refrained from using the system for systematic
  experimental work to favor reproducibility and due to limited available
  resources.
  In more detail, the AMD Rome systems consist of 2 CPUs each having a
  price of about 7.000~EUR, which makes such systems not widely
  available and limits access for many researchers.
  Furthermore, the system is part of a TOP500 cluster~\cite{taurus},
  requiring detailed billing of system usage measured by CPU hours.
  Since SAT solvers are highly memory intensive, one often favors 
  employing the memory caches lines well. %
  We observed a different number of solved instances when used 8 or
  32 cores.
  One might take the number of available memory channels to be 
  very conservative. In our case, each CPU consists of
  128 logical cores (64 physical cores) but only 16 memory channels.
  However, this results in high overbilling of cluster resources,
  usually counted by reserved logical cores.
  At the same time, the Rome CPU has a base-frequency of only 2.0GHz
  where one would expect slower solving times.
  However, the system has a faster cacheline and much larger L1, L2,
  and L3 caches. This might result in faster solving or  yield
  unexpected results as some solvers are sophistically engineered for
  memory cachelines~\cite{FichteMantheySchidler20a}.
  In order to favor reproducibility, we decided on a standard and
  widely used system.
\end{FIXED}%
Thus, Team~\teamB's system contained two Intel Xeon Silver 4112
CPUs (Skylake architecture) of 2.60GHz base-frequency equipped with
128GB RAM. We ran the experiments at the \sFIXED{base-frequency}.
Since the Netra X1 from 2000 was equipped with 2GB and the NetBSD
allowed to still run all source code based solvers, even the very
modern ones, the Sun system serves as a point of reference.

\subsection{The Final Stage: Experimental Setting and Limitations}
We compare wall clock time and the number of timeouts. However, we avoid
IO access on the CPU solvers whenever possible, i.e., we load
instances into the RAM if a network file system is involved and store
uncompressed instances. We set a timeout of 900 seconds and limited
available RAM to 512MB per instance and solver.  We also tested for
some solvers with resident set size restricted to 1GB RAM and observed
only a minimal difference. Since Intel hardware around 2002 rarely
had more than 512MB RAM available, we went for the 512MB setup.
We follow
standard guidelines for benchmarking~\cite{KouweAndriesseBos18a}.
We set and enforce resource limits by the tool
\textit{runsolver}~\cite{Roussel11a}.
\begin{FIXED}%
  We are aware that runsolver suffers from sampling-based issues.
  Measuring resources results in immediately expired
  information, and enforcing limits might not affect if the used
  RSS (resident set size) exceeds the intended maximum limit only in
  sudden resource spikes~\cite{BeyerLoweWendler15,BeyerLoweWendler19}.
  We have no strong indicators that this is the case in our setting
  and hence still favor runsolver over a significantly more complex
  system setup using cgroups.
  We force performance governors to the respective base-frequency for
  modern CPUs~\cite{SchoneEtAl19a}, disable hyperthreading, and
  enforce the process that handles the solver invocation to run on at
  most the number of cores that equals the number of memory channels.
  We explicitly enable transparent huge pages systems with modern CPUs
  and large caches, which is also the default with SAT competitions
  that run on StarExec~\cite{StumpSutcliffeTinelli14}.
  Note that the Pentium~3 system has non-ECC memory, which might cause
  memory errors and affect results~\cite{SchroederPinheiroWeber09}.
\end{FIXED}%

\begin{table}
\centering
\begin{tabular}{lllrr}
\toprule
group &                          Solver & Year/Generation &  HW99 &  HW19 \\
\midrule
      &             \solver{MapleSat19} &            2019 &    \multirow{3}{*}{\rotatebox[origin=c]{90}{{\tiny Team~\teamA}}~~}
                                                              \textbf{72} &   187 \\
      &                \solver{CaDiCal} &            2019 &   \textbf{100} &   184 \\
      &                \solver{Glucose} &            2016 &   \textbf{105} &   188 \\
  \midrule
      &                  \solver{vbest} &                 &   124 &   198 \\
      &                    sum &                 &   277 &   559 \\
      &                    avg (\%) &                 &    46 &    93 \\
  \midrule
  \midrule
      &               \solver{Forklift} &            2003 &    43 &   \multirow{3}{*}{\rotatebox[origin=c]{90}{{\tiny Team~\teamB}}~~} 
                                                                    \textbf{105} \\
      &              \solver{siege\_v3} &            2003 &    43 &    \textbf{96} \\
      &                \solver{Jerusat} &            2003 &    38 &    \textbf{71} \\
  \midrule
      &                  \solver{vbest} &                 &    53 &   134 \\
      &                    sum &                 &   124 &   272 \\
      &                    avg (\%) &                 &    21 &    45 \\
  \midrule
  \midrule
  \multirow{6}{*}{\rotatebox[origin=c]{90}{Knuth}}      &               \solver{SAT13+12} &          CDCL+P &    36 &   111 \\
      &                  \solver{SAT13} &            CDCL &    30 &    98 \\
      &               \solver{SAT11+12} &            LH+P &    16 &    23 \\
      &                  \solver{SAT11} &              LH &    16 &    19 \\
      &               \solver{SAT10+12} &          DPLL+P &    12 &    12 \\
      &                  \solver{SAT10} &            DPLL &     6 &     6 \\
  \midrule
  \midrule
  \multirow{8}{*}{\rotatebox[origin=c]{90}{Other Solvers}}      &              \solver{Lingeling} &            2019 &    67 &   178 \\
      &  \solver{Lingeling-aqw-27d9fd4} &            2013 &    85 &   186 \\
      &          \solver{Lingeling-276} &            2011 &    80 &   177 \\
      &                \solver{MiniSat} &            2008 &    83 &   173 \\
      &              \solver{siege\_v4} &            2004 &    50 &   116 \\
      &              \solver{siege\_v1} &            2003 &    34 &    85 \\
      &                \solver{Compsat} &            2003 &    47 &    92 \\
      &             \solver{BerkMin561} &            2003 &    36 &    88 \\
      &             \solver{zChaff(32)} &            2001 &    40 &    62 \\
      &                   \solver{sato} &            2000 &    17 &    19 \\
      &                   \solver{satz} &            1998 &     7 &     9 \\
\bottomrule
\end{tabular}

\caption{Overview of the number of solved instances for the various solvers on 
  our old and new hardware. \texttt{HW99} represents the number of solved instances on the
  old hardware. \texttt{HW19} represents the number of solved instances on
  the new hardware.
  \solver{vbest} represents best virtual solvers, 
    which are  virtual solvers that we obtain by taking all instances that have been 
    solved by the solvers considered in the group listed above.
}
\label{tab:results}
\end{table}

\begin{figure}[t]
\centering
\vspace{-3em}
\resizebox{0.8\columnwidth}{!}{%
  \includegraphics{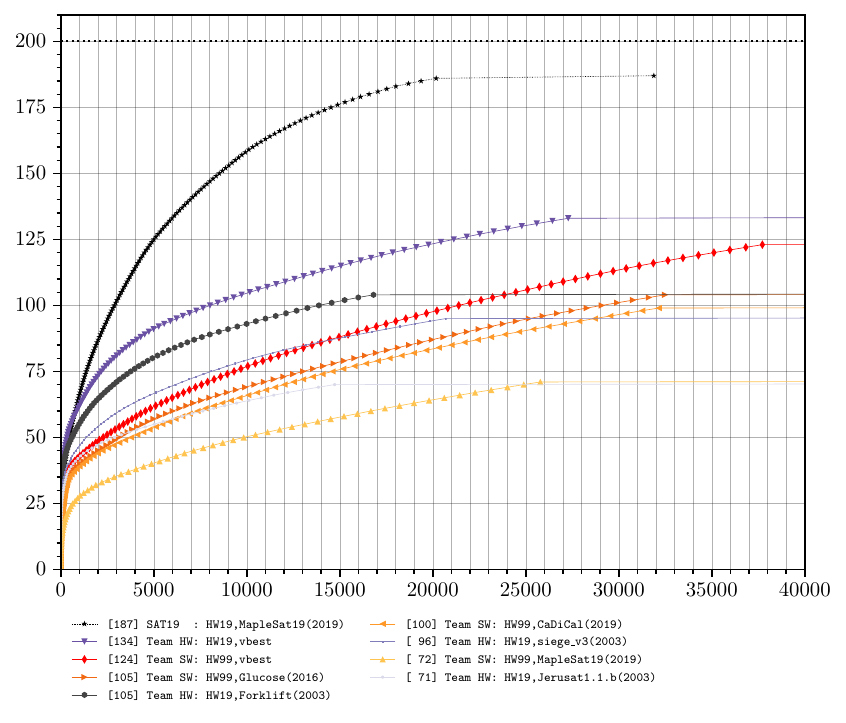}
}
\caption{%
  \sFIXED{%
    Total runtime and cumulated solved instances for each 
  considered solver. 
  The dashed line indicates the total number of instances in our
  benchmark set.
  The maximum possible total runtime for each solver would be 
  180.000s (we observed a maximum of ~140.000s). 
  However, we limited the x-axis to 40.000s 
  as (when ignoring timeouts) 
  each solver successfully solved all its solvable instances within a
  total runtime of less than 40.000s.
  Each line illustrates a solver and the maximum number of 
  instances it successfully solved if allowed to run 
  for x seconds.
  The last data point represents, if shown, the total runtime 
  of a solver.
  The x-axis refers to the total runtime. The y-axis depicts
  the maximum number of instances solved within this runtime. 
  \solver{vbest} refers to the best virtual  solver,~i.e., we take
  the union over the solved instances for each team and consider the
  minimum for each instance. In the legend, $[X]$ refers to the number
  of~$X$ solved instances. \texttt{HW19} refers to the new hardware,
  and \texttt{HW99} refers to the old hardware. \texttt{SAT19} refers
  to a modern solver on modern hardware, which one can consider as
  a potential baseline.
  }%
}
\label{fig:cdf}
\end{figure}

\section{The Trophies}
\label{sec:results}
Table~\ref{tab:results} gives an overview on the number of solved
instances for each solver and the two hardware generations.
Figure~\ref{fig:cdf} illustrates the runtime of the selected
solvers and hardware as a CDF-like plot.
Our results and gathered source codes are all publicly
available~\cite{FichteHecherSzeider20a}.
Note that we report only on the two Intel-based hardware generations
in this table. The results on the Ultra Sparc~IIe system look very
similar, usually, a few more instances were solved. 
Detailed data can be found in the supplemental material.

\subsection{Results}%

\begin{FIXED}
  Before we start discussing the results, we would like to point out
  that we report fewer solved instances here than in the preliminary
  version (cf.~\cite{FichteHecherSzeider20a}). In the present work, we
  enforce the CPU governors to the base frequency, which is lower than
  the maximum turbo frequency (as used in the preliminary work). Using
  the CPU at its maximum possible frequency requires a sophisticated
  system setup ensuring that the frequency is reproducible and the
  same for all solvers at all times. Setting the maximum possible
  frequency is not enough as the CPU might not run at its maximum
  turbo frequency and automatically throttle down due to thermal
  restrictions. A setup running at the maximum possible frequency
  requires observing the active CPU frequency constantly and checking
  whether it agrees with the maximum turbo frequency. In order to
  favor reproducibility, we opted against running the systems at their
  maximum frequency.

When we consider the number of solved instances on the hardware from
2019, \solver{MapleSat19} solves 195 instances. 
Recall that Team~\teamB consists of the old solvers on modern
hardware. They solve 105 instances (\solver{Forklift}), 96 instances
(\solver{siege\_v3}), and 71 instances (\solver{Jerusat}).  On
average, they solve about~91 instances (45.5\% of the instances) at a
standard deviation of about 18.
However, the virtual best solver (\solver{vbest}) for Team~\teamB solves
134 instances,~i.e., about 67\% of the instances. The virtual best
solver is the virtual solver that we obtain from taking the union over
the solved instances by all three solvers and keeping the instance
with the best solved runtime.
The Team~\teamA consists of the new solvers on old hardware. They solve
72 instances (\solver{MapleSat19}), 100 instances (\solver{CaDiCal}),
and 105 instances (\solver{Glucose}). On average, they solved about 92
instances (46.0 \% of the instances) with a standard deviation of
18. Their virtual best solver (\solver{vbest}) solves 124
instances,~i.e., about 62.0\% of the instances.
When considering the solvers \solver{MapleSat19},
\solver{CaDiCal}, and \solver{Glucose} on modern hardware, they solve
 186 instances on average with a very low standard deviation of~2
instances.
When considering the results on the solvers \solver{Forklift},
\solver{siege\_v3}, and \solver{Jerusat}) on old hardware, they solve on
average about~41 instances (21.5\% of the instances) at a standard
deviation of about 3.
\end{FIXED}

\subsection{Discussion of the Results}
\subsubsection{Comparing the Teams%
}
\begin{FIXED}
The solver \solver{Glucose}, which is a very stable solver from 2016,
solves the highest number of instances on the new hardware.
\end{FIXED}
We are not surprised that neither Team~\teamA nor Team~\teamB or their
virtual best solver gets anywhere close to this result.
In view of Table~\ref{tab:results} and Figure~\ref{fig:cdf}, there
are plenty of ways to compare the two teams. One can carry out (i)~an
individual comparison by the best (\solver{vbest}), worst, or average solver, or even
consider the individual solvers in direct comparison to each other, 
but one could also (ii)~consider the virtual best solver for
each team.
If we choose Method~(i) and individually compare the solvers, Team~\teamA
clearly wins for the measure best, worst, or average solver.
We can also do one-by-one comparison and compare the solvers from each
team individually with the solvers from the other team. Then, we take
the number of solved instances for each solver~X from Team~\teamA against
each solver~Y from Team~\teamB, and we give X a point if it solves more
instances than~Y or give a point to Y in the opposite case.
\begin{FIXED}
  Then, \solver{Glucose} obtains 2 points (because it solves more
  instances than \solver{siege\_v3} and \solver{Jerusat});
  \solver{CaDiCal} obtains 2 points, and \solver{MapleSat19} obtains 0
  points, which totals 4 points for Team~\teamA. In comparison,
  Team~\teamB receives 2 points for \solver{Forklift}, 1 point for
  \solver{siege\_v3}, and 0 points for \solver{Jerusat}, which totals 3
  points.
Hence, Team~\teamA also wins.
Nevertheless, if we consider the virtual best solvers, Team~\teamB
performs better than Team~\teamA.

\end{FIXED}

\subsubsection{Notable Observations}
\label{sec:results:observations}
\begin{FIXED}
We found it surprising that the winner from the 2019 SAT Race
(\solver{MapleSat19/\allowbreak HW99}) solves fewer instances than two
solvers from Team \teamA (\solver{Forklift/\allowbreak HW19}, 
\solver{siege\_v3/\allowbreak HW19}) of Team~\teamB.
This seems surprising to us, and we currently do not have a good
explanation of why \solver{MapleSat19} solves so few instances on the old
hardware, namely 28 instances less than \solver{CaDiCal} and 33
instances less than \solver{Glucose}. 
One might suspect that \solver{MapleSat19} runs out of memory as it
stores hashes even for learnt clauses that have been deleted. However,
we did not observe this on our instances. %
We observed a similar behavior with the latest implementation of
\solver{Lingeling} but not with \solver{CaDiCal}, which also implements inprocessing
techniques.
Hence, we suspect that the advanced data structures in the
solvers supporting large caches, the learning and restarting policy,
and strong tuning towards modern hardware might be contributing
factors.

We found it interesting that the old solvers \solver{Forklift},
\solver{siege\_v3}, and \solver{Jerusat} still solve a considerable
number of instances on the new hardware. In particular, the solver
\solver{Forklift} and \solver{siege\_v3} seem to benefit substantially
from the new hardware, while \solver{Jerusat} gains only a small
benefit from the new hardware.
Since \solver{siege\_v3} already implements cache-aware algorithms
that might highly benefit from fast and large memory
cachelines, the behavior is not entirely surprising~\cite{Ryan02}.
In contrast, solvers that use pre-CDCL techniques such as Knuth's
\solver{SAT10} solver, \solver{sato}, and \solver{satz} could not
benefit from new hardware.
When we consider Knuth's implementations, it is particularly
remarkable that only solvers with modern techniques (CDCL) benefit
from new hardware.

\end{FIXED}

\subsection{Summary}
When reviewing the results, we believe that our test-setting
revealed that both Team~\teamA and~\teamB perform in a similar range.
If we compare individually, Team~\teamA wins, which is also well visible in
the CDF-like plot in Figure~\ref{fig:cdf}. However, if we consider virtual
best solvers, Team~\teamB performs equally well.
This leaves us with the conclusion that the last decades have brought
enormous technological progress and innovation for \SAThy solving. 
The two main factors, (i)~\emph{hardware advancement} and
(ii)~\emph{algorithm advancement}, both have a considerable influence.
\begin{FIXED}
  Results on the solvers by Donald Knuth, which represent different
  epochs of SAT solving, illustrate quite well that solvers benefit
  only from modern hardware after advanced algorithmic techniques
  (CDCL) have been integrated.
  Still, modern SAT solvers advanced far beyond efficient
  solving. Proof logging and automated proof verification allow for
  employing modern SAT solvers even to establish mathematical
  proofs~\cite{HeuleKullmann17}.
\end{FIXED}

\section{Conclusion}
We compare the impact of hardware  and
algorithm advancement on a genuine problem, namely the propositional
satisfiability problem (SAT).
We describe the decisions and challenges from a
thought experiment to an actual experiment between old solvers and new
solvers on new and old hardware with a time difference of about
two decades.
Our experiment's outcome confirms that modern algorithms  have a
strong influence on  the  solvers' performance, even when they
run on old hardware. Nonetheless, solving significantly profits from
technological advancement in hardware development. There is no
clear winner between Team~\teamA (new solvers on old hardware)
vs. Team~\teamB (old solvers on new hardware) in our time leap challenge.
Overall, both teams perform in a similar range with a slight advantage
for Team~\teamA (new solvers on old hardware), which leads us to
conclude that both hardware and software advances in science and
industry have a mutual influence on modern solving. Hence, algorithm
advancements are at least as important for the field of \SAThy solving
as hardware advancement.
Further, algorithm engineering becomes of more importance.

During our research, we noticed that long term reproducibility highly
depends on available source code or static binaries with few
dependencies.  Further, it turned out to be helpful if the setup of a solver
requires few additional system tools and few dependencies on external
libraries.
The dependencies within the operating system and source codes usually
were not the problem as architectural dependencies would 
forbid to run the solvers.
From our archaeological investigations, we suggest avoiding any
external system for the setup for future long term experiments,~i.e.,
tight dependencies on kernel versions or software containers, such as
Docker. Still, one uniform shared system for the entire
  community, such as the SAT Heritage project, might prove
  helpful~\cite{AudemardPauleveSimon20a} if implemented by
  competition organizers.
Further, we think that public data libraries would help 
understand long term advancements, not just source code repositories
of private companies or university webpages.

One could post an open call and repeat the experiment with any
solver. However, we believe that this would probably challenge
developers of modern solvers to optimize their implementation for old
hardware, which would result in a distorted picture for old solvers.
Hence, we do not primarily intend to repeat the experiments in
  the near future~\cite{Munroe19a}.

We hope that our work stimulates research for others to set up a
time leap challenge in their fields, such as stochastic
SAT-solvers, CSP-solvers, MaxSAT-solvers, and ILP-solvers.

\section*{Acknowledgments}

We would like to thank several colleagues.  Dave Mitchell and João
Marques-Silva supported us with source codes from their old disks or
mailboxes.
Uwe Pretzsch and Siegmar Schoene helped to organize old hardware and
Toni Pisjak maintained the modern benchmarking system, which we used.
\begin{FIXED}
  We thank Armin Biere for his comments on the validity of the
  produced results and correctness of early solvers and Mate Soos for
  providing us with details on the memory behavior of the solver
  \solver{MapleSat19}.
\end{FIXED}

%

%
%
%
%
%
%
%
%
%
%
%
%
%
%
%
%
%
%
%
%

%
%
%
%
%
%
%
%
%
%
%
%
%
%
%
%
%
%
%
  
%
%
%
%
%
%
%
%
%
%
%
%
%
%
%
%
%
%
%
%
%
%
%
%
%
%
%
%
%
%
%
%
%
%
%
%
%
%
%
%
%
%
%

%

%
%
%

%
%
%
%
%
%
%
%
%
%
%
%
%
%
%
%
%
%
%
%
%
%
%
%
%
%
%
%
%

\end{document}